# Joint Prediction of Remaining Useful Life and Failure Type of Train Wheelsets: A Multi-task Learning Approach

Weixin Wang
Department of Industrial and Systems Engineering
University at Buffalo, The State University of New York
1775 Wiehle Ave #340, Reston, VA 20190
Phone: (703)945-0286
Email: wangweixin23@gmail.com

Qing He[1]
Department of Civil, Structural and Environmental Engineering and
Department of Industrial and Systems Engineering
University at Buffalo, The State University of New York
313 Bell Hall, Buffalo, NY 14260
Phone: (716)645-3470
Email: qinghe@buffalo.edu

Yu Cui
Department of Civil, Structural and Environmental Engineering
University at Buffalo, The State University of New York
204 Ketter Hall, Buffalo, NY 14260
Phone: (716) 645-4351
Email: ycui4@buffalo.edu

Zhiguo Li
IBM T J Watson Research Center
1101 Route 134 Kitchawan Rd,
Yorktown Heights, NY 10598
Phone: (914) 945-3000
Email: lizh@us.ibm.com

---

[1] Corresponding author

# ABSTRACT

The failures of train wheels account for disruptions of train operations and even a large portion of train derailments. Remaining useful life (RUL) of a wheelset measures the how soon the next failure will arrive, and the failure type reveals how severe the failure will be. RUL prediction is a regression task, whereas failure type is a classification task. In this paper, we propose a multi-task learning approach to jointly accomplish these two tasks by using a common input space to achieve more desirable results. We develop a convex optimization formulation to integrate both least square loss and the negative maximum likelihood of logistic regression, and model the joint sparsity as the L2/L1 norm of the model parameters to couple feature selection across tasks. The experiment results show that our method outperforms the single task learning method by 3% in prediction accuracy.

# 1. INTRODUCTION

Wheelsets are the top rolling stock maintenance item in North America. Transportation Technology Center, Inc (TTCI) showed that wheelset replacement costs $828 million annually, while the number is still increasing (Cummings 2012). Wheelset defects build by a variety of causes, from rolling contact fatigue to out-of-roundness of the wheel. In real-world operations, a defective rail wheel generates a large amount of issues, including high impact loads, bearing failures, and even severe accidents (Railway-technical.com 2016).

Wheelset maintenance is essential to prevent railcars from set-outs, In-service-failure (ISF), and train derailments. Moreover, some researchers proved that re-profiling wheelsets after certain mileage would nearly double the wheel service life thus minimizing total life cycle costs (Braghin et al. 2006). Therefore, it is critical to estimate the remaining time before the next occurrence of a failure given the current and historical wheel conditions (Jie, 2018). We define a random variable:

$$Z(t) = T_f - t \mid T_f > t$$

where $T_f$ denotes the random variable of failure time, and suppose that the wheel has survived until time *t*, as depicted in Figure **1**.

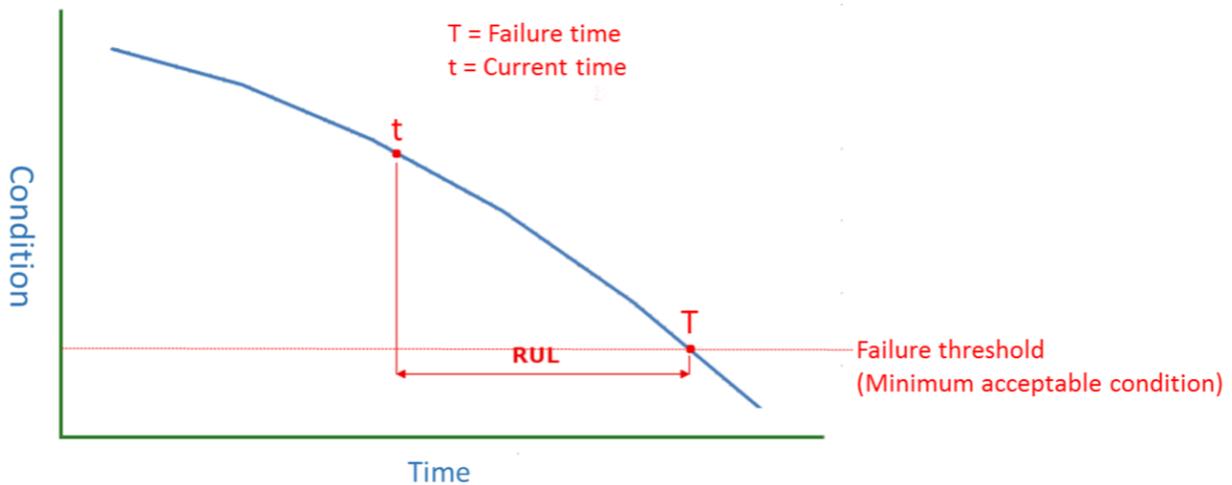

**Figure 2** Illustration of the concept of remaining useful life (RUL)

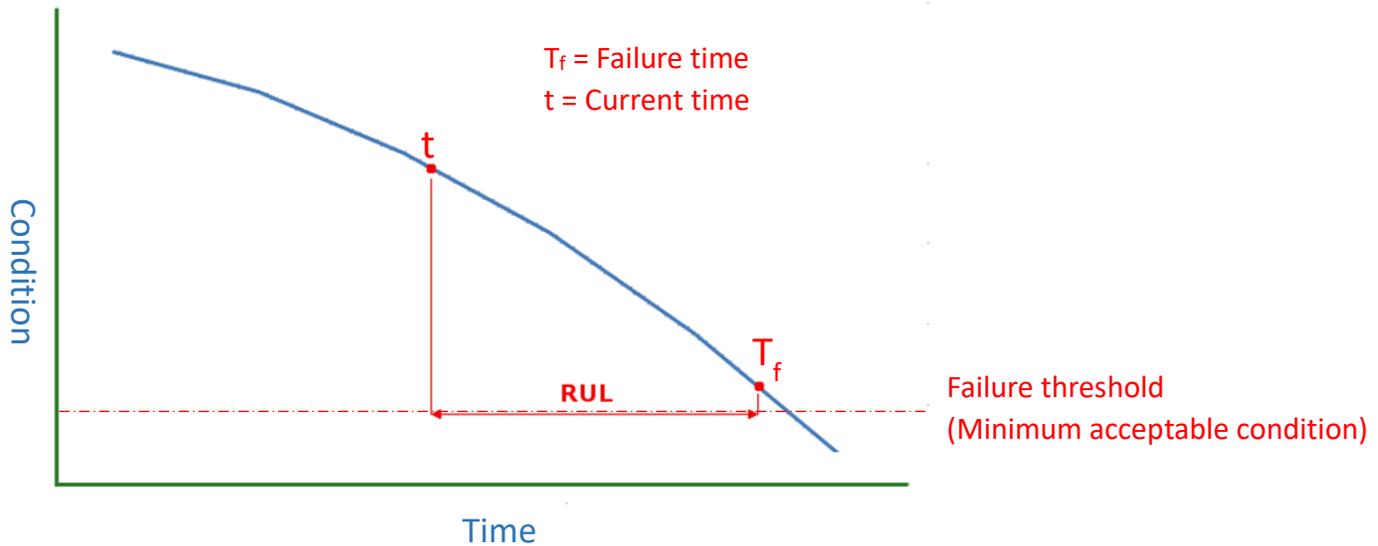

In the literature of reliability engineering, $Z(t)$ is also known as remaining useful life (RUL). RUL, also called remaining service life, residual life or remnant life, is nowadays in fashion, both in theory and applications (Banjevic 2009, Cheng 2018). The concept of the RUL is upon individual explanations, regarding the word "useful". A recent study conducted a complete survey of statistical data-driven approaches for RUL estimation models, which can be further classified into two broad types of models, that is, the RUL estimation models based on the directly observed state processes, and those cannot be observed directly (Si et al. 2011). This paper only considers failure prediction for a wheelset. Existing maintenance practice always replaces or re-profiles both wheels in one axle at the same time, since one wheel degenerates rapidly once the other wheel is defective.

In additional to RUL, accurate prediction of the failure type is also critical. Defective wheelsets are treated with two repair action types: replacement with a new wheelset (N), or wheelset turning (T). Some defective wheelsets that violate the Federal Railroad Administration (FRA) track safety standards have to be replaced. Others that do not violate the standards, however, will eventually degrade to type N defective wheelsets if they are not fixed in time. The correct anticipation of failure type can benefit preventive maintenance, reduce maintenance cost, and decrease derailment risk. To better define the problem, we define the class attribute $C_i$ as the failure type of each wheelset $i$ using following equation:

$$C_i(t) = \begin{cases} 0 \\ 1 \end{cases} \quad \text{when t (current time)} \geq T_f \text{ (failure time)}$$

where $T_f$ and $t$ represent the failure time and current time, respectively. 0 denotes that the wheelset only needs to be turned or reprofiled (T) and 1 the wheelset has to be replaced with a new pair (N). If t is lees than $T_f$, we are not able to define a failure since wheel damage has not reached failure threshold (Figure1).

The objective of this paper is to develop a prognostic machine learning model to concurrently predict RUL and failure type of wheelsets in a freight car with readings from wayside detectors. Competing risk models are employed to analyze the failure probability under different failure types, different car kinds, and wheel sizes. Further, a multi-task learning method is applied for common feature selection, followed by prediction with Support Vector Regression (SVR) and Support Vector Machine (SVM). The model is expected to assist railway companies in making decisions about rail vehicle operation and condition-based maintenance (Palo et al. 2012), by providing further information on the causes of wheel defects and degradation.

The contributions of this paper lie on the following aspects:
1) As the first attempt, this paper analyzes the rail wheelset failure time under two different failure types with the competing risk model.
2) As another first attempt of leveraging multi-task learning in rail transportation, this paper develops a convex optimization formulation to integrate both least square loss and the negative maximum likelihood of logistic regression, and model the joint sparsity as the L2/L1 norm of the model parameters to couple feature selection across tasks.
3) This paper develops a gradient descent algorithm to solve proposed convex optimization model.

The rest of the paper is organized as follows. Section 2 briefly reviews the predictive maintenance, railway failure prediction, and multi-task learning. Section 3 discusses the data collection and initial analysis. Section 4 focuses on feature selection with multi-task learning. A new methodology is proposed to combine least square loss and the negative maximum likelihood of logistic regression and solved by the gradient descent method. Following the methodology, Section 5 presents the experiment results based on the data depicted in Section 3. The conclusions of proposed methodology and the future work are given in Section 6.

## 2. LITERATURE REVIEW

As discussed in Section 1, wheel failures cost significantly to railway operations. The first question to be answered is how to identify a failure event. Palo et al. (2012) stated that condition monitoring of railway vehicles is mainly performed using wheel impact load detectors (WILD) and truck performance detectors (TPD). Yang and Létourneau (2005) declared a failure event when a wheel vertical impact measurement exceeds 140 kips (A kip is an Imperial unit of force. It equals 1000 pounds-force) or a computed value greater than 170 kips. Li and He (2015) used bad order data directly, consisting of order dates and types. Once a bad order is issued, the car is scheduled to be checked or repaired in the workshop. Hajibabai et al. (2012) categorized an instance of train stop as a failure; otherwise, with no train stop nor repair record, it was categorized as a non-failed wheel. Stratman et al. (2007) stated that Current AAR's wheel impact load limit for wayside detectors, which is 90 kips, is no longer reliable, and since many catastrophic wheel failures occurred below the 90 kips.

Failure prediction is the next step of failure identification. In failure prediction, there are three major methods: (1) develop criteria, indices and pertinent thresholds for wheel failure, (2) statistical learning and data mining based methods (Z. Li and He 2015) and (3) degradation model based failure prediction. Ekberg et al. (2002) defined three indices to quantify fatigue impact. They also set corresponding pertinent thresholds. If one or more inequalities are fulfilled, fatigue is predicted to occur. They noticed that all the conservative approximation occurred at the same lateral position. A more precise estimation can be made by studying the lateral spread of fatigue impact from a given load. Yang and Létourneau (2005) proposed a Multiple Classifier System (MCS) capable of predicting 97% of wheel failures while maintaining a reasonable false alert rate (8%). The model only gave the probability of failure but cannot tell the occurrence time of the failure. Li and He (2015) developed a Random Forests based methodology to assess the current health and predict RUL of both trucks and wheels of a railcar. They also compared the efficiency of three types of detectors. But they did not consider the failure types and the following repair actions. Actually, the costs of reprofiling a wheel and replacing a wheel are totally different. Hajibabai et al. (2012) developed a logistic regression model to classify wheel failures with a classification accuracy of 90% and 10% false alarm rate. However, they only take into account wheels within 30 days of train stop as "bad wheels", which means that the model cannot predict the failure times that are greater than 30 days.

Degradation models are also prevailing analytic tools for RUL estimation (Ye and Xie 2015). Two broad categories of degradation models are general path models and stochastic process models. General path models are easy to use but difficult to capture dynamic feature of complex system. Stochastic models for example, Wiener process, gamma process and Markov process are on the other way around (Ye and Xie 2015). Ye and Chen proposed a new class of random effects model for the Wiener process model (Ye, Chen et al. 2015). They used it to analyze a dataset of fatigue crack growth and Hard Disk Drives (HDD) wear data, which demonstrated the new degradation-stress relationship. Si and Hu developed a two-state continuous-time homogeneous Markov process to approximate the switches between the working state and storage state (Si, Hu et al. 2014). However, the accuracy of the residual storage life is affected by the selection of hyper-parameters (Prashanth L.A, 2016). Si also proposed a nonlinear degradation model with Kalman filtering to estimate RUL, which outperforms the conventional linear Wiener process model according to experiment on battery data (Si 2015).

The multi-task learning approach provides a good solution for prediction of both RUL and failure type. Different from the conventional single-task feature selection, the multi-task feature selection simultaneously selects a common feature subset relevant to all tasks (Zhang et al. 2012).

Our proposed algorithm shares some similarities with recent work in (X. Yang et al. 2009) where they deal with heterogeneous tasks including both continuous and discrete outputs from a common set of input variables (Kun Lin, 2018). Two main differences are that their formulation uses L-∞ regularization and, in our formulation, we implement L2/L1 norm, a combination of L1 and L2 norm that proves to be a better approach to combine tasks and ensure that common features will be selected across the group (Evgeniou and Pontil 2007). It is well-known that using L1 norm leads to sparse solutions (Xinyan Zhao, 2018), which allow some components of the learned vector to be zero so that we can select those features that matter. For most large underdetermined systems of linear equations, the minimal L1-norm solution is also the sparsest solution. Note that this L1/L2 regularization scheme reduces to the L1 regularization in the single-task case, and can thus be seen as an extension of it where instead of summing the absolute values of coefficients associated to features, we sum the Euclidean norms of coefficient blocks. The L2-norm is used here as a measure of magnitude, and one could also use $L_p$-norms (Obozinski et al. 2006).

## 3. DATA DESCRIPTION and PREPARATION

We collected our data from one of Class I railroads in North America. The entire datasets include more than 2-year maintenance data, bad order data, mileage data and WILD data from Jan 2010 to Mar 2012. The following subsections will summarize data fields, data cleaning, and processing of every dataset. Also, a statistical summary of RUL is included, and competing risks analysis is applied to each failure type.

**3.1 Data presentation**

Typically defective wheels generate high impact load on the track. WILD is built into the track to detect defective wheels, weighing each wheel several times when the wheel passes by a detector in a certain distance (Lechowicz and Hunt 1999). WILD uses strain-gauge-based technologies to measure the performance of a railcar in a dynamic mode (H. Li et al. 2014). The strain gauges quantify the force applied to the rail through a mathematical relationship between the applied load and the strain caused to the rail web or rail foot (Stratman et al. 2007). Once a train is detected, WILD generates different levels of data include: train data, equipment data, truck data and wheel data. Some important features are shown in Table 1.

Once a failure is diagnosed by a bad order (which defines the failure time in this paper), the faulty equipment will be scheduled to visit workshop depending on the severity level of failure. There are dozens of reasons that may contribute a bad order, such as thin brakes and warm bearings. In this paper, only wheel bad orders are considered.

Maintenance data, containing actual repair actions, was collected from railcar workshops. For wheel repairs, technicians usually replace or turn the defective wheel in pairs, depending on damage intensity. The corresponding failures are labeled with two categories: New (N) and Turn (T). New, simply means a severe failure that leads to new wheelset replacement, whereas turn represents a failure that involves re-profiling or turning.

Only equipment with more than one repair record will remain in the final table since the first repair type (or failure type) will be used to predict the following repair type (or failure type). Without previous maintenance record, we can not determine start of repair cycle and remaining useful life. Records with RUL less than 60 days were excluded from this study. When RUL is very short, it is more reasonable to apply failure detection other than RUL prediction (Z. Li and He 2015). Moreover, to address relatively near failures, data records with RUL greater than 180 days were removed from the final dataset as well. Furthermore, to study how axle side (left or right) influences on RUL and failure type, data with only one side records was also omitted from the raw data. . Since road condition influencing on left and right axle side is not even. Accordingly, left and right axle sides may need different maintenance, which might be repaired or totally changed (failure type). Although it has more than 10 types of car in the original dataset, some car types only account for a very small proportion. We merged those car types to one called miscellaneous type, denote as M. This results in only four types: gondola, hopper, hopper without cover and miscellaneous. After data extraction and cleaning, the final dataset consists of 2459 observations and 110 features.

## 3.2 Competing risks analysis for failure times and failure types

The maintenance data measures the time span from the origin until the occurrence of one type of failure. If several types of failures occur, a model describing progression to each of these competing risks is needed (Putter et al. 2007). In this paper, two failure risks are addressed including "New" and "Turn", which involve different maintenance costs. We apply the competing risk model to analyze the failure time that is defined by the difference between the first repair time and the next bad order time (Scrucca et al. 2007).

Figure 2 presents the cumulative failure probability within certain days for different combinations of car kinds and wheel size. G, H, L and M represent four different car kinds: gondola (G), hopper (H), hopper without cover (L) and miscellaneous (M), respectively. 33 and 36 are wheel sizes in inch. N and T indicate the first repair type, either new replacement (N) or turning (T). As one can see from Figure 2(a) and Figure 2(b), no matter what car kind or wheel size it is, failure type leading to "N" occurs much more frequently than type "T". For every car type (see Figure 2(a)), the failure probability for "N" is very close to each other among different car kinds, so is the "T" risk. However, "N" risk is apparently much higher than "Turn". The probability of replacing a new wheelset is around 80% within two years, whereas the probability of turning the wheelset is 20% in the same period. Therefore, it is found that in the most of the time, wheel failure cannot be fixed without replacing a new wheelset. Such expenditure spent on wheelset replacement is anticipated to decrease if more preventive maintenance is applied. Furthermore, when we consider wheel size (Figure 2(b)), the 33-inch wheel shows the highest failure probability, which indicates that existing maintenance needs to be improved when applied on 33-inch wheels.

**Table 1 WILD data and data connections**

**Train Data**
Each train has at least one locomotive and several equipment. Those equipment may belong to different companies.

| Attribute | Description |
|---|---|
| DET_ID | Unique identifier for an electronic detector |
| CAR_CNT | Count of rail cars in a train. |
| MAX_PK_KP | Max peak wheel load reading in kips |
| MAX_KP_RA | Max ratio between average & peak kips |
| LOCO_ID | Locomotive name |
| MAX_HUNT | The max train car truck hunting index |
| TRN_KIND | Kind of train |

**Equipment Data**
Each equipment has at least two trucks. Or more than 20 trucks in rare cases.

| Attribute | Description |
|---|---|
| DET_ID | Unique identifier for an electronic detector |
| EQP_INIT | Equipment initial |
| AXLE_CNT | Axel amout |
| TRCK_CNT | Truck amount in one equipment |
| LST_MAINT | User id that changed this row |

**Truck Data**
Each truck has two axles. Those trucks could be of different types.

| Attribute | Description |
|---|---|
| DET_ID | Unique identifier for an electronic detector |
| TRCK_SEQ | The sequence of a truck on a car |
| WGT_TONS | The weight of a truck recorded in tons |
| HUNT_IDX | The truck hunting index for each truck |
| RMDTN_TS | Indicates timestamp |

**Wheel Data**
Each axle has two wheels, left and right.

| Attribute | Description |
|---|---|
| DET_ID | Unique identifier for an electronic detector |
| AXLE_NBR | Identify an axle on specific wheel |
| AXLE_SIDE | R=right, l=left |
| AVG_KIPS | Average load reading in kips for a wheel |
| PEAK_KIPS | Peak load reading kips for wheel |
| LAT_KIPS | Ave lateral load reading kips for a wheel |
| LAT_PEAK | Peak lateral load reading kips for a wheel |

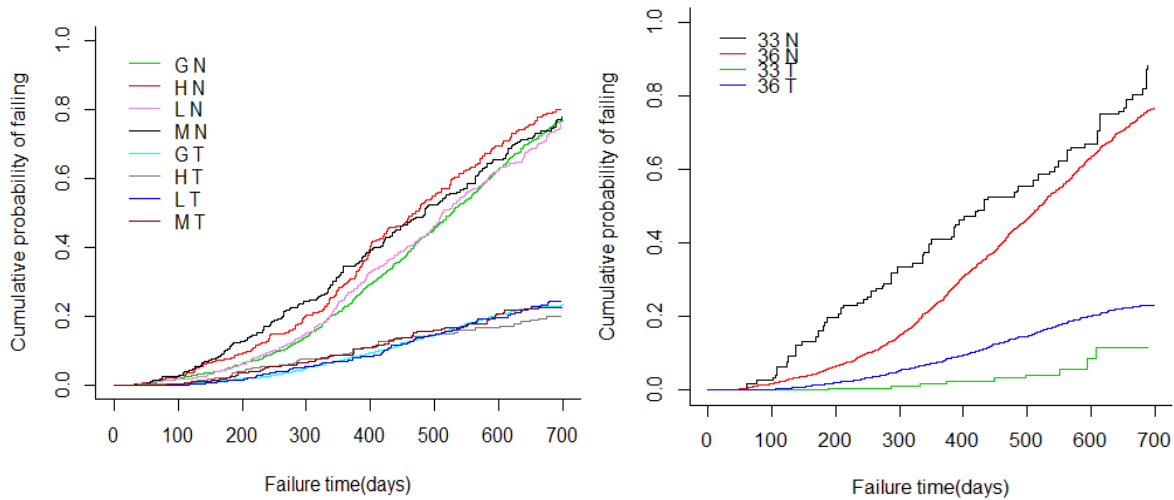

**Figure 2(a)**        **Figure 2(b)**

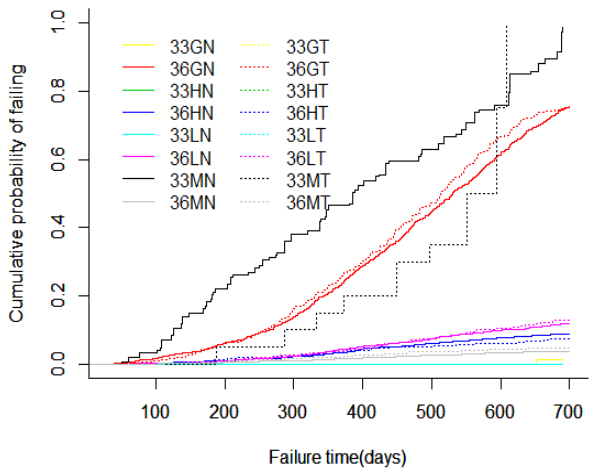

**Figure 2(c)**

**Figure 2 The results of competing risk analysis: (a) Competing risk for different car kinds; (b) Competing risk for different wheel size; (c) Competing risk for car type and wheel size**

We further investigate the failure probability under categories classified by both car type and wheel size, as shown in Figure 2(c). All solid lines represent type "N" risk and dash lines represent type "T" risk. It shows a similar pattern as the previous two figures that given certain number of days, the probability of wheelset replacement ("N") is higher than the one of turned wheels ("T"). One exception is that car kind G with a 36-inch wheel, represented by a red line, has high failure probability under both risks. Another exception is observed for 33-inch miscellaneous car. Its failure probability of "T" increases quickly and even exceeds its "N" risk after 600 days. One possible reason is that "M" is a mixed category that contains car kinds with a long maintenance cycle.

## 4. METHODOLOGY and ALGORITHM

Our methodology consists of two major parts: Multi-task feature selection (MTFS) and Multi-modal support vector machine (SVM) and support vector regression (SVR). In this paper, we conduct two tasks, including the regression task of RUL and the classification task of corresponding failure types. MFTS aims to select common feature for these two related tasks. The underlying assumption is that prediction of RUL will provide useful information for classification of failure types and vice versa. Based on jointly selected features; we implement SVM/SVR to conduct the prediction and classification for these tasks, respectively. Figure 3 illustrates the flow chart of the proposed methodology.

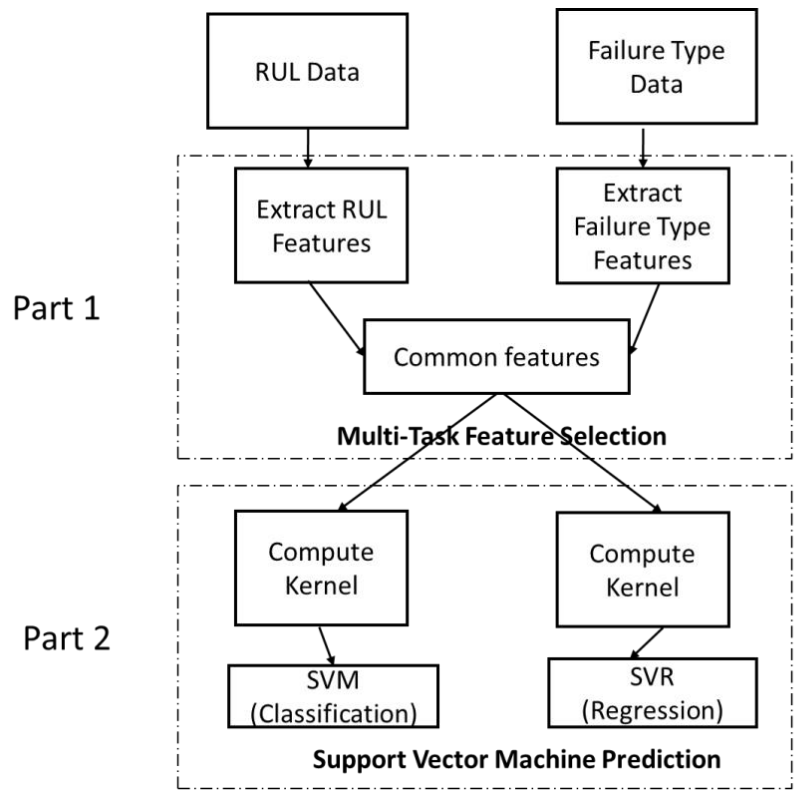

**Figure 3 Procedure of proposed methodology**

## 4.1 Multi-task feature selection (MTFS)

Different from the conventional single-task feature selection, multi-task feature selection simultaneously selects a common feature subset relevant to all tasks. MTFS is especially important for the diagnosis of failure time and type, since they are both essentially determined by the same underlying attributes, i.e., car kind, wheel load, wheel size, etc. (Zhang et al. 2012).

The learning algorithm simultaneously learns all the tasks through two alternating steps. The first step is independently learning the parameters of the regression/classification tasks. The second step consists of learning, in an unsupervised way, a low-dimensional representation for these task parameters, which we show to be equivalent tolerance common features across the tasks. The number of common features learned is controlled by the regularization parameters (X. Yang et al. 2009).

### 4.1.1 Regression task

RUL, denoted as $y_i^r$ in this paper, is defined as number of days between current meaturement date and the following bad order date for $i$th observation. Obviously, $y_i^r$ is a continuous variable that we can model it as a linear regression where:

$$y_i^r = a_0 + \sum_{j=1}^{m} x_{ij} a_j + \epsilon, i = 1, \ldots, n$$

Suppose we have $n$ observations, while each observation has $m$ features. $x_{ij}$ represents $i^{th}$ observation's $j^{th}$ feature. And $a_j$ represents coefficient for $j^{th}$ feature, while $a_0$ is the intercept and $\epsilon$ the residual.

Accordingly, least square is adopted as the loss function of the regression task, which is formulated as follows:

$$L_r = \sum_{i=1}^{n}(y_i^r - a_0 - \sum_{j=1}^{m} x_{ij} a_j)^2$$

### 4.1.2 Classification task

Wheelset repair can be either replacement or turning. We assume that a wheelset goes back to original brand new state after either repair activity is applied. If the wheelset is badly damaged so that maintenance technicians have to replace the whole wheelset, we denote its corresponding failure type for replacement as 1, otherwise, denote it as 0. Let $y_i^c$, a binary variable, represents the failure type for $i$th observation. We use the logistic regression to model it (Czepiel 2002).

$$Prob(y_i^c) = \frac{1}{1 + \exp(-b_0 - \sum_{j=1}^{m} x_{ij} b_j)}, i = 1, \dots, n$$

Classification task shares the same inputs with regression task. Identically, $b_j$ represents classification coefficient for $j^{th}$ feature and $b_0$ the intercept. To estimate parameters $b_j$, one can minimize the negative log-likelihood (X. Yang et al. 2009) given as below:

$$L_C = -\sum_{i=1}^{n} \left\{ y_i^c \left( b_0 + \sum_{j=1}^{m} x_{ij} b_j \right) + \log \left[ 1 + \exp(b_0 + \sum_{j=1}^{m} x_{ij} b_j) \right] \right\}$$

### 4.1.3 Regularization

Figure 4 illustrates the details of regularization processes. The key point of MTFS is to first apply L2-norm for both tasks. As a result, weights corresponding to the $j^{th}$ feature across multiple tasks, are forced to be grouped together and tend to be selected jointly as a group (Zhang et al. 2012). Furthermore, we also applied L1 regularization to obtain a sparse solution of MTFS results, in which the weights of groups of features are forced to be zero. We do not penalize $a_0$ and $b_0$ since it is not desirable for the model to depend on the mean of the *y* vector (Schmidt 2005).

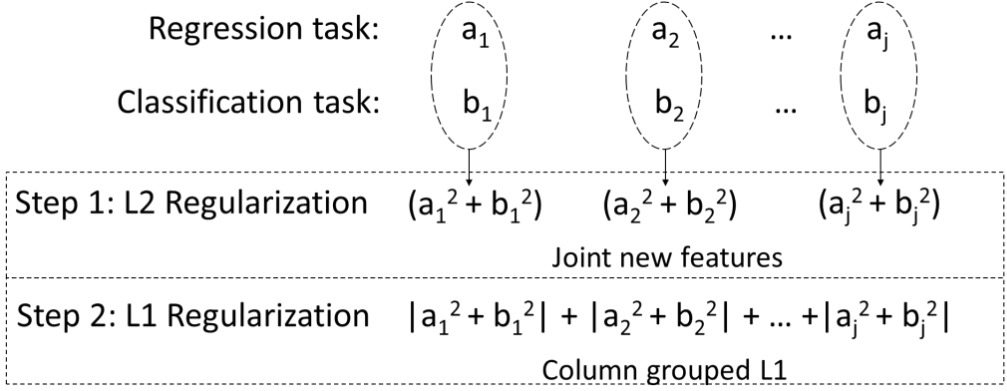

**Figure 4 Diagram for L2/L1 regularization**

Let $L_n$ be the regularization penalty, shown as follows:

$$L_n = \sum_{j=1}^{m}(a_j^2 + b_j^2)$$

### 4.1.4 Objective function

Based on previously defined three components, the objective function can be easily constructed as follows:

$$\text{Minimize: } L = \theta L_r + L_c + \lambda L_n$$

$$\text{Subject to: } L_r = \sum_{i=1}^{n}(y_i^r + a_0 + \sum_{j=1}^{m} x_{ij} a_j)^2$$

$$L_c = -\sum_{i=1}^{n}\left\{y_i^c\left(b_0 + \sum_{j=1}^{m} x_{ij} b_j\right) + \log\left[1 + \exp\left(b_0 + \sum_{j=1}^{m} x_{ij} b_j\right)\right]\right\}$$

$$L_n = \sum_{j=1}^{m}(a_j^2 + b_j^2)$$

where λ is a regularization parameter that determines the sparsity level, and it controls the number of selected features, i.e., the "group sparsity". Because of the characteristic of 'group sparsity', the solution of MTFS results in a weight matrix whose elements in some columns (groups) are all zeros. For feature selection, we only keep those features with non-zero weights. Note that at preprocessing stage, we perform a common feature normalization step, i.e., subtracting the mean and then dividing the standard deviation (of all training subjects) for each feature.

## 4.2 Algorithm

### 4.2.1 Gradient descent algorithm

Several different methods can be used to solve the proposed convex optimization formulation, such as gradient descent, steepest descent, Newton's method and interior method. Second-order derivatives methods have fast convergence to approach a global minimum of convex objective functions, but they also involve computing Hessian matrix and its inverse matrix, which most likely would be infeasible in the high-dimensional setting. Considering this, we adopt gradient descent method (Boyd and Vandenberghe 2004). This method iteratively updates each element of the parameter vector once at a time, using a closed-form update equation given all of the other elements. Besides, the loss functions of linear regression and logistic regression have different forms. The gradient descent method optimizes their original loss function without any transformation so that it is more intuitive to see how the two heterogeneous tasks affect each other. The method can be briefly elaborated as the following steps:

Given an initial guess the weight matrix $w = w_{initial}$

Repeat

**Step 1** Compute gradient $g(a_j, b_j)$;

**Step 2** Update $w$ by $w_{new} = w - \gamma * g(a_j, b_j)$;

**Step 3** Checking stop criterion: quit if mean *(abs ($w_{new}$ - w) /abs (w)) < tolerance*;

**Step 4** Update $w$ by $w = w_{new}$.

Where $\gamma$ is the learning rate. It controls how big the step is when updating the parameters. If $\gamma$ is very large, it corresponds to very aggressive gradient procedure. If $\gamma$ is very small, it corresponds to small steps. To find a proper $\gamma$, which making sufficiently fast converge progress, we employed an approche called a line search. In this paper, we choosed final value for γ is $10^{-4}$. Figure 5(c) shows the gradient descent after 100 steps with above γ value.

### 4.2.2 Objective function in matrix format

$$L_r = (Y_r - XA)^t \cdot (Y_r - XA)$$

$$L_c = -Y_c^t \cdot (XB) + \sum_{j=1}^{m} \log(1 + \exp(XB))$$

$$L_n = A^t \cdot A + B^t \cdot B$$

- Input data $X$: nxm matrix
- Output for regression (RUL) $Y_r$ : nx1 vector
- Output for classification (failure type) $Y_c$ : nx1 vector
- Loss function for regression $L_r$: nx1 vector
- Loss function for classification $L_c$: nx1 vector
- Loss function for classification $L_n$: nxn vectorCoefficient vector for regression $A$: nx1 vector
- Coefficient vector for classification $B$: nx1 vector

### 4.2.3 Proof of convexity

Before applying gradient descent method, we need to prove that the objective function is convex. If it is not a convex function, there is no optimal solution that is guaranteed. Since each coefficient is a decision variable, it will be more convenient to solve the problem with its matrix format (Bazaraa et al. 2013). The convexity of a function can be proved by its Hessian matrix, which is a square matrix of second-order partial derivatives of a scalar-valued function. If the Hessian matrix is positive semi-definite on the interior of the convex set, it is a convex function (Ruszczyński 2006). The proof can be found in Appendix.

**4.3 SVM and SVR prediction**

Support Vector Machine (SVM) prediction, is the process of learning to separate samples into different classes (Shin, Lee et al. 2005). SVM performs classification by finding the hyperplane that maximizes the margin between the two classes (Melgani and Bruzzone 2004). The vectors that define the hyperplane are the support vectors. To define an optimal hyperplane we need to maximize the width of the margin (*w*) (Huang and Wang 2006).

The Support Vector Regression (SVR) uses the same principles as the SVM for classification, while maintaining the differences between estimated values and real values under $\varepsilon$ precision since the output is a real number (Lin, Cheng et al. 2006). However, the main idea is always the same: to minimize error, individualizing the hyperplane that maximizes the margin, keeping in mind that part of the error is tolerated (Taghavifar and Mardani 2014).

**5. EXPERIMENT RESULTS**

[This experiment consists of two major parts: 1) Multi-task feature selection (MTFS) and 2) support vector regression (SVR) and support vector machine (SVM). Several evaluation indicators, such as MAPE, precision, and recall, are used to measure the efficiency of the proposed model.] This experiment consists of two major parts: feature selection and prediction by SVM. This whole paper emphasizes on Multi-task feature selection (MTFS). We choose single feature selection and random forest feature selection as benchmark models. To measure the efficiency of the proposed model, we run support vector regression (SVR) and support vector machine (SVM) to perform prediction, based on the features selected by 3 different model. Several evaluation indicators, such as MAPE, precision, and recall, are used to measure the efficiency of the proposed model.

**5. 1 MTFS and random forest**

**5.1.1 MTFS experiment procedure and results**

After data extraction and cleaning, the final dataset consists of 2459 observations and 110 features (M= 2459, N=110). We have demonstrated that our algorithm is able to find the global optimum in Section 4. It guarantees that for a fixed dataset, our result is the combination of most significant features. The main idea of MTFS is to select significant features through adjusting the value of parameter $\lambda$. Different $\lambda$ results in different subsets of original feature pool. The procedure of MTFS can be described as follows.

**Step1**: Apply gradient algorithm and adjust $\lambda$ and $\theta$;
**Step2**: Decide selection criteria and get common features;
**Step3**: Conduct cross-validation and get an optimal solution.

The above steps show details of the procedure of MTFS. In the first step, the value of $\lambda$ and $\theta$ vary from 0.001 to 1000. When $\lambda$ increases, it reduces the value of coefficients towards 0 gradually. Moreover, because of the fixed stopping criteria, the iteration could stop when it reaches the tolerance level. Figure 5(a) and Figure 5(b) show how $\lambda$ affects the objective value and iteration numbers. As one can see, when $\lambda$ is larger, the objective value tends to converge faster.

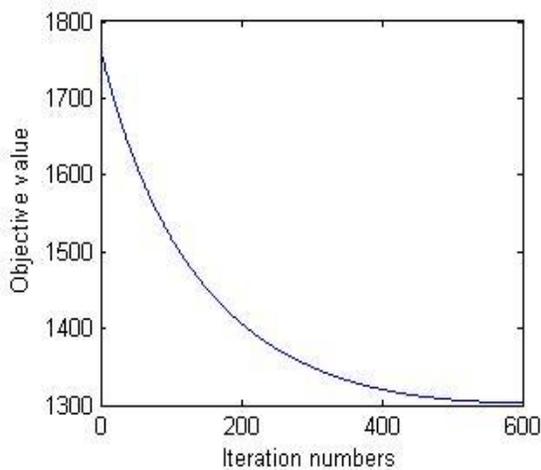
**Figure 5(a)**

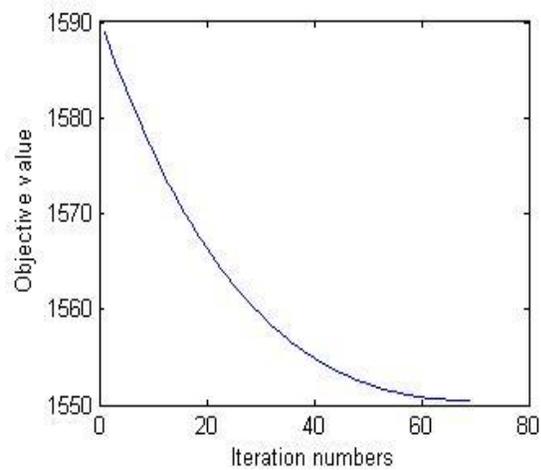
**Figure 5(b)**

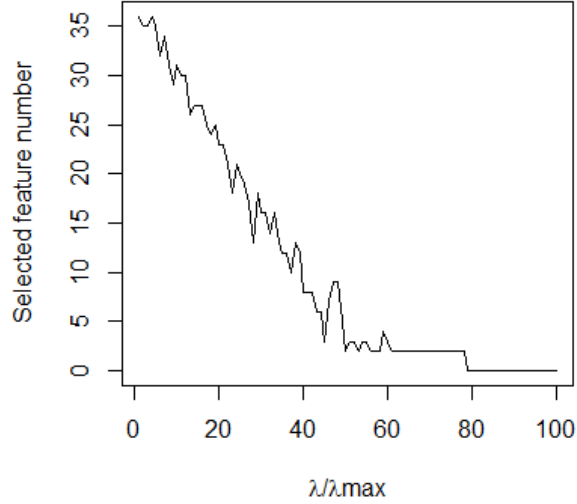

**Figure 5(c)**

**Figure 5 The computation results of MTFS: (a) Objective value converge curve when λ = 1; (b) Objective value converge curve when λ = 10; (c) Relation between the number of selected features and the percentage of λ/ λ$_{max}$**

For different tasks, it is reasonable to set different selection criteria since the loss functions of linear regression and logistic regression have different forms. For example, features are considered to be significant if the corresponding coefficients are greater than 0.01 in regression task. However, the key point of MTFS is to obtain common features. Within the feature subset, which is demonstrated as significant through regression task, we need to remove those features that are non-significant in classification tasks. Figure 5(c) displays how different λ affects the number of selected features, which decreases with an increasing λ. However, in the range [0, 60%] for the ratio λ/ λ$_{max}$, the number of selected features goes up and down, which means some noise exists. In the range [80%, 100%], when λ is very big, all coefficients are forced to 0 thus no feature is selected.

In step 3, we select a solution with a proper number of features, and we validate the solution with cross-validation. Overfitting generally occurs if a model contains too many features. In this case, the model describes random error or noise instead of the underlying relationship. Or the model could become under-fitting when it cannot capture the underlying trend of the data if too few features are included. Hundreds of experiments are conducted to search a suitable $\lambda$ and the corresponding number of selected features. Some results are presented in Table 2. Compared to $L_r$ (least square) $L_c$ varies dramatically. The reason is that maximum likelihood is very sensitive since it includes exponential function.

**Table 2 Part of experiment results**

| Trial # | $\theta$ | $\lambda/\lambda_{max}$* | L | $L_c$ | $\lambda *L_n$ | $\theta*L_r$ | Iteration# | Selected Feature # |
|---|---|---|---|---|---|---|---|---|
| 1 | 100 | 0.01 | 434.8894 | 184.9594 | 67.56367 | 182.3664 | 8553 | 36 |
| 2 | 100 | 0.12 | 1132.455 | 196.0116 | 754.0382 | 182.4052 | 498 | 31 |
| 3 | 100 | 0.28 | 1253.554 | 425.6094 | 644.5609 | 183.3835 | 237 | 23 |
| **4** | **100** | **0.42** | **1371.195** | **547.0092** | **640.1895** | **183.9963** | **213** | **18** |
| 5 | 100 | 0.41 | 1359.133 | 544.0439 | 631.1071 | 183.9816 | 174 | 8 |
| 6 | 100 | 0.47 | 1402.299 | 584.0818 | 634.038 | 184.179 | 131 | 3 |
| 7 | 100 | 0.53 | 1451.84 | 609.2766 | 658.2805 | 184.2824 | 125 | 3 |
| 8 | 100 | 0.68 | 1512.169 | 699.1607 | 628.3714 | 184.6374 | 115 | 2 |
| 9 | 100 | 0.74 | 1548.669 | 714.7982 | 649.1702 | 184.7007 | 98 | 0 |
| 10 | 100 | 0.93 | 1593.573 | 802.3792 | 606.1568 | 185.0373 | 85 | 0 |

* $\lambda_{max}$: maximum value of $\lambda$, which is 100.

In the end, as shown in Table 2, the highlighted trial (#4) is selected as the final solution of MTFS, which contains 18 selected features. Cause:

As depicted in Table 3. Those jointly selected features show the consistency between each other. For example, TRAIN_TYPE. C, indicating whether the train transports coal, is selected together with CAR_KIND.G, representing if a railcar is a gondola, designed for ore product.

**Table 3 Selected features and descriptions by MTFS**

| LABELS | EXPLANATION |
|---|---|
| **WHL_A_KIPS.R** | Average load reading taken in kips for an individual car wheel |
| **N_WHL_A_KIPS.R** | Normalized WHL_A_KIPS for right wheel |
| **TTRCK_WGT_TONS** | Weight of a train car truck recorded in tons |
| **EDR_EQP_SPD** | Speed of an equipment |
| **CAR_CNT** | Count of railcars in a train. |
| EDR_HMDTY_PCT | Humidity percentage |
| EDR_WIND_DIR | Wind direction |
| VNDR_TRN_TYP.F | A general type of train as assigned by a vendor. F= Freight |
| **MAX_PK_KP** | Max peak wheel load reading in kips |
| TTRCK_MAX_TONS | Maximum weight of a train car truck recorded |
| VNDR_LD_CD.L | Load status for a car or train. 0=Empty 1=loaded |
| VNDR_LD_D.M | Empty status for a car or train 0=loaded 1=Empty |
| **MAX_HUNT** | The max train car truck hunting index |
| TRN_KIND.C | Kind of train. C=COAL |
| **FIR_APPLD_JCD** | The last repair applied job code |
| CAR_KIND.G | Car kind. G=GONDOLA |
| SIZEE.36 | Wheel size. 36=36 inches |
| FIR_REP | Last repair type |

* Bold features are also selected by random forest feature selection.

5.1.2 Random forest results

To verify the effectiveness of MTFS, we use random forest model to conduct feature selection and serve as a benchmark model. In Li and He (2015), random forest effectively handled missing data in wayside detector readings . It also has significant advantage to our experiment such as: ranked by feature's relevance, less sensitive for outliers and faster to train data than SVM. Figure 6 illustrates the entire feature selection process, in which 14 features are found to obtain the minimal RMSE. Table 4 displays all 14 features selected by random forest. It also ranks all the variables by relevance. Compare to the results of MTFS, random forest produces 7 common features, which are bolded in Table 4.

**Table 4 Selected features and descriptions by Random forest**

| LABELS | EXPLANATION |
|---|---|
| CAR_AXLE_CNT | Number of axles on all cars in a train |
| **N_WHL_A_KIPS.R** | Normalized WHL_A_KIPS for right wheel |
| **CAR_CNT** | Count of railcars in a train. |
| **WHL_A_KIPS.R** | Average load reading taken in kips for an individual car wheel |
| **TTRCK_WGT_TONS** | Weight of a truck recorder in tons |
| **N_WHL_A_KIPS.R** | Normalized WHL_A_KIPS for right wheel |
| WHL_AVG_KIPS.L | Average downward load readings taken in kips for wheel |
| WHL_PEAK_KIPS.R | Peak downward load readings taken in kips for wheel |
| EDR_EQP_SPD | Speed of an equipment at instance |
| EDR_AMB_TMP | Ambient air temperature at a specific electronic detector reading |
| **FIR_APPLD_JCD** | The last repair applied job code |
| MAX_KIPS_RATIO | Max ratio between average and peak kips wheel readings |
| TTRCK_LGTH_IN | Length of a truck recorded in inches |
| **MAX_HUNT_IDX** | Max train hunting index when a train pass by an electronic detector |

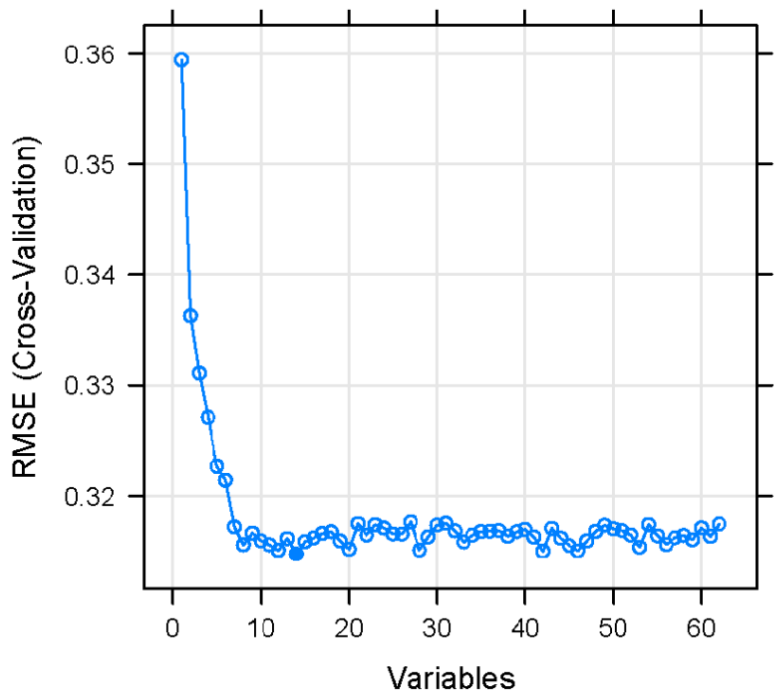

**Figure 6  Automatically selected and ranked features by random forest**

## 5.2 SVM Result

To get a reliable result, we use k-fold cross-validation to testify our prediction results based on jointly selected features. The advantage of this method over repeated random sub-sampling is that all observations are used for both training and validation, and each observation is used for validation exactly once. 5-fold cross-validation is used in this paper, which means 80% data is used as training data and 20% as test data at each iteration.

In this paper, SVM model is trained to distinguish between "New" (N) and "Turn" (T). To visualize the performance of SVM classification result, receiver operating characteristic (ROC) curve is introduced as Figure 7(a). The curve is created by plotting the true positive rate against the false positive rate at various threshold settings. The closer the curve follows the left-hand border and then the top boundary of the ROC space, the more accurate the test. This ROC curve is above the 45-degree diagonal but not very far from it, which means there is still room for improvement.

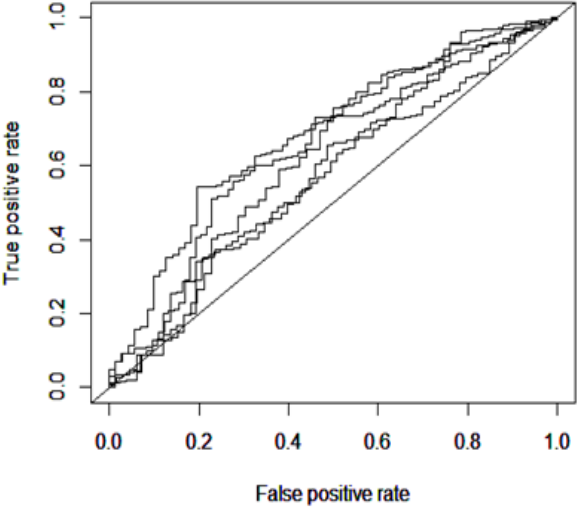

**Figure 7(a)**

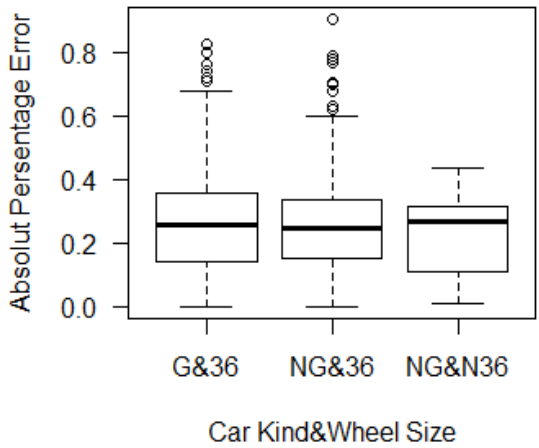 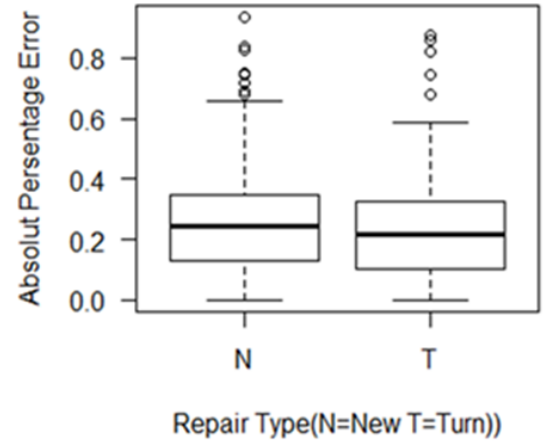

Figure 7(b)                                                        Figure 7(c)

**Figure 7** The results of the proposed model: (a) ROC curve for SVM; (b) APE of different car kind& wheel size for SVR; (c) APE of repair types for SVR. Note: G= gondola, NG= Non-gondola, and 36= 36-inch wheel, N36= non-36-inch wheel

For the regression task, we evaluate its performance through absolute percentage error (APE) and mean absolute percentage error (MAPE) of predicted RUL. Since the car kind of gondola and 36-inch wheel dominate in the dataset, we divide the dataset into 4 subsets, which is gondola &36 inches, gondola &non-36 inches, non-gondola & non-36 inches. In our dataset, there is no gondola car with 33 inches wheel size involved. Figure 7(b) shows the results of APE for different subsets. From the perspective of car kind and wheel size, we can conclude that it's relatively more robust to predict category in non-gondola&non-36 inches since the range of its APE is smaller, compared to other categories. In addition, there is no evident difference between APE of type "New" and type "Turn" Figure 7(c). It is found that the SVR model works equally well for both failure types.

## 5.3 Cross-validation results

Cross-validation and its repeated test are applied in this paper. For regression task, we take MAPE to measure the accuracy of the regression. For classification task, we calculate precision and recall based on confusion matrix to evaluate classification result. Precision and recall are defined as follows,

$$\text{Precision} = TP/(TP + FP)$$

$$\text{Recall} = TP/(TP + FN)$$

where TP is True positive; TN is True negative; FP is False positive; FN is False negative.

**Table 5 The cross-validation result among three different feature selection methods**

|  |  | RUL prediction with SVR | Failure type classification with SVM | |
|---|---|---|---|---|
|  |  | MAPE | Precision | Recall |
| Method | Single Task Feature Selection | 0.280 | 0.800 | 0.788 |
|  | Random Forest Feature Selection | 0.260 | 0.789 | 0.805 |
|  | MTFS | 0.252 | 0.809 | 0.824 |

We compare the results of MTFS with the results of single task feature selection and Random Forest feature selection, shown in Table 5. Single task model selects features for regression and classification respectively, by only L1 regularization. Compare to random forest feature selection, Single task model shows higher precision but lower recall and higher MAPE. Therefore, the performance of Random Forest slightly improves single task feature selection method. Among this three method, MTFS outperforms the other two methods. Compared to single task model, the MAPE of RUL prediction and the recall of failure type classification are improved by 3%. Compared to Random Forest, the MAPE of RUL prediction is reduced by 1% and the precision and recall of failure type classification are improved by 2%. The results from cross-validation validate that multi-task learning works well in wheel failure prediction that jointly considers the failure time and the failure type.

## 6. CONCLUSIONS and FUTURE WORK

### 6.1 Conclusions

In this paper, we propose a multi-task learning method to jointly select common features for prediction of wheelsets Remaining Useful Life (RUL) and failure types. Such a method combines linear regression loss, logistic regression loss and L2/L1 regularization for multi-task feature selection (MTFS). In our experiments, wheel measurement data from WILD is incorporated with bad order data and repair data to a comprehensive table, which is further divided to training dataset and test dataset to perform cross-validation. We demonstrate that using L2/L1 regularizations not only selects features but also leads to "group sparsity", which identifies the input variables that are commonly relevant to multiple tasks.

In this study, 18 features are selected and considered as inputs for later prediction part. Besides, those jointly selected features have shown the consistency between each other.

The prediction consists of two components, SVM for failure type classification and SVR for RUL prediction. The prediction results show that both the recall of classification and the MAPE of RUL prediction is reduced by around 3%, compared to the one of single-task learning. It is also found that the wheelsets with car kind gondola and 36-inch wheel are relatively easier to be predicted by comparing across different categories.

Moreover, competing risk analysis revealed that wheelsets tend to fall into severe failure that requires replacement ("New"). Most of time, mechanics tend to replace the defective wheelsets with a new wheelset, no matter what car kind or wheel size is. Only a small portion of wheel failure is rectified by turning the wheels or re-profiling the wheels ("Turn").

In summary, our experimental results show that our proposed multi-task learning method can effectively predict remaining useful life and failure type of wheelsets concurrently.

## 6.2 Future work

We can extend the proposed method to second-order derivatives method to prepare for large-scale datasets and more tasks. Also, some other machine learning methods could be developed to increase prediction accuracy for remaining useful life. One another future research direction is to apply proposed method to optimize maintenance scheduling (Ye, Chen et al. 2015)

**Appendix:**

**The proof of the convexity of Hessian matrix of *L***

**Proof**: We denote objective function *L*'s Hessian matrix of as H. Hessian matrix is a square matrix of second-order partial derivatives of a scalar-valued function.

In the second derivative test for determining extrema of objective function L(A,B), H is given by:

$$H = \begin{bmatrix} \dfrac{\partial^2 L}{\partial A^2} & \dfrac{\partial^2 L}{\partial A\, \partial B} \\ \dfrac{\partial^2 L}{\partial B\, \partial A} & \dfrac{\partial^2 L}{\partial B^2} \end{bmatrix}$$

For better understanding, we described H as a 2nx2n square matrix, consisting of 4 partitions: P1 to P4, shown as below.

$$H = \begin{bmatrix} P_1 & P_2 \\ P_3 & P_4 \end{bmatrix}$$

$$P_1 = \frac{\partial^2(L)}{\partial^2(A)} = \theta * 2 \cdot X' \cdot X + \lambda * 2I$$

$$P_2 = P_3 = \frac{\partial^2(L)}{\partial(A)\partial(B)} = 0$$

$$P_4 = \frac{\partial^2(L)}{\partial^2(B)} = X' \cdot (1 - \pi) \cdot \pi' \cdot X + \lambda * 2I$$

Thus, H can be transformed to block diagonal matrix D.

$$D = \begin{bmatrix} P_1 & 0 \\ 0 & P_4 \end{bmatrix}$$

For D, it is straightforward to see that for every non-zero column vector $z$ of n real numbers, $z^T*D*z$ is greater or equal to 0. Therefore, H is a convex function.